**Dual-path convolutional neural network using micro-FTIR imaging to predict breast cancer subtypes and biomarkers levels: estrogen receptor, progesterone receptor, HER2 and Ki67.**

Matheus del-Valle[a], Emerson Soares Bernardes[b], Denise Maria Zezell[c,*]

[a] Lasers and Applications Center, Energy and Nuclear Research Institute, University of São Paulo, São Paulo, Brazil

[b] Radiopharmacy Center, Energy and Nuclear Research Institute, University of São Paulo, São Paulo, Brazil

* Corresponding author.

E-mail addresses: matheusdv@gmail.com (M. del-Valle), emerson.bernardes@gmail.com (E.S. Bernardes), zezell@usp.br (D.M. Zezell).

**Abstract**

Breast cancer molecular subtypes classification plays an import role to sort patients with divergent prognosis. The biomarkers used are Estrogen Receptor (ER), Progesterone Receptor (PR), HER2, and Ki67. Based on these biomarkers expression levels, subtypes are classified as Luminal A (LA), Luminal B (LB), HER2 subtype, and Triple-Negative Breast Cancer (TNBC). Immunohistochemistry is used to classify subtypes, although interlaboratory and interobserver variations can affect its accuracy, besides being a time-consuming technique. The Fourier transform infrared micro-spectroscopy may be coupled with deep learning for cancer evaluation, where there is still a lack of studies for subtypes and biomarker levels prediction. This study presents a novel 2D deep learning approach to achieve these predictions. Sixty micro-FTIR images of 320x320 pixels were collected from a human breast biopsies microarray. Data were clustered by K-means, preprocessed and 32x32 patches were generated using a fully automated approach. CaReNet-V2, a novel convolutional neural network, was developed to classify breast cancer (CA) vs adjacent tissue (AT) and molecular subtypes, and to predict biomarkers level. The clustering method enabled to remove non-tissue pixels. Test accuracies for CA vs AT and subtype were above 0.84. The model enabled the prediction of ER, PR, and HER2 levels, where borderline values showed lower performance (minimum accuracy of 0.54). Ki67 percentage regression demonstrated a mean error of 3.6%. Thus, CaReNet-V2 is a potential technique for breast cancer biopsies evaluation, standing out as a screening analysis technique and helping to prioritize patients.

## 1. Introduction

Female breast cancer is the most incident cancer with 11.7%, or 2.3 million, of new cases in 2020, aside from 6.9%, or 690 thousand, of deaths [1]. The classification for the breast cancer can follow different parameters, as stage, grade, and molecular subtypes. Molecular subtypes classification plays an import role in the breast cancer treatment, sorting patients with divergent prognosis and helping to select an appropriate and specific therapy [2]. Subtypes are defined using the expression levels of Ki67 biomarker and three hormone receptors: estrogen receptor (ER), progesterone receptor (PR) and human epidermal growth factor receptor 2 (HER2). The four subtypes and their usual treatments are [3,4]:

- Luminal A (ER and/or PR positive, HER negative, Ki67 low) – endocrine therapy.
- Luminal B (ER and/or PR positive, HER variable, Ki67 high) – endocrine therapy and chemotherapy; if HER2 positive, anti-HER2 therapy may also be used.
- HER2 subtype (ER and PR negative, HER positive, Ki67 usually high) – chemotherapy and anti-HER2 therapy

- Triple-negative (ER, PR and HER negative, Ki67 usually high) – chemotherapy.

While histology and immunohistochemistry techniques are widely used to classify breast cancer subtypes, interlaboratory and interobserver variations can affect the accuracy of these methods [5], besides being time-consuming and laborious techniques, where the lack of pathologists in most countries aggravates the situation [6]. Fourier Transform Infrared (FTIR) spectroscopy has been studied as a further cancer evaluation technique in the past years, not only to overcome the variations, but also to provide additional information [7,8].

With the consolidation of Fourier Transform Infrared micro-spectroscopy (micro-FTIR) imaging, which provides thousands of spectra in a single acquisition, and the need of automated tools, machine learning approaches stood out as powerful tools for many diagnostics, including cancer classification [9,10]. The deep learning subarea of machine learning has become one of the most important tool for artificial intelligence [11]. Researches have applied deep learning in several spectral domains [12], including biospectroscopy/biospectral imaging [13] and vibrational spectroscopy [14]. To the date, there is no study using FTIR and deep learning for breast cancer subtypes assessment, being limited to malignant vs benign diagnosis using ATR-FTIR single spectra acquisition [15] or blood serum [16], and morphological comparison with chemical histology techniques [17].

In this way, there is still a lack of studies using recent deep learning techniques to evaluate breast cancer, their molecular subtypes and biomarkers expression levels using micro-FTIR hyperspectral images. A complete automated analysis tool could provide extra information for the pathology report and act as screening technique, speeding up the patient assessment and prioritization process.

## 2. Material and Methods

*2.1. Dataset*

A total of 60 cores from the BR804b (Biomax, Inc, USA) breast cancer microarray was imaged. Histological sections of 8 µm were formalin fixed and paraffin embedded (FFPE) in calcium fluoride ($CaF_2$) slides (Crystran, UK). The company also provided the Ground Truth (GT) labeling of receptors and Ki67 expression levels using immunohistochemistry (IHC). Molecular subtypes were classified in accordance with St. Gallen International Expert Consensus guidelines [3,4]. **Table 1** presents the distribution of the dataset acquired.

**Table 1**

Dataset distribution regarding each label. Type with AT (adjacent tissue) and CA (cancer) classes; Subtype with LA (luminal A), LB (luminal B), HER2 and TNBC (triple-negative breast cancer); ER (estrogen receptor), PR (progesterone receptor), and HER2 receptors expression levels; Ki67 percentage levels. HER2 1+ and 2+, and other Ki67 levels were not considered due to small quantity (only one core).

| Label | Class | Quantity |
|---|---|---|
| Type | AT | 30 |
|  | CA | 30 |
| Subtype | LA | 8 |
|  | LB | 8 |
|  | HER2 | 7 |
|  | TNBC | 7 |
| ER | – | 11 |
|  | + | 3 |
|  | ++ | 3 |
|  | +++ | 13 |
| PR | – | 15 |
|  | + | 3 |
|  | ++ | 2 |
|  | +++ | 10 |
| HER2 | 0 | 16 |
|  | 3+ | 11 |
| Ki67 | 5% | 12 |
|  | 10% | 6 |
|  | 20% | 4 |
|  | 30% | 3 |

Hyperspectral images mosaics of 320x320 were acquired using a Cary Series 600 system (Agilent Technologies, USA) with a focal plane array (FPA) detector of 32x32 and spatial resolution of 5.5 µm, resulting in a total of 6,144,000 raw spectra for the 60 cores. The full equipment range was collected, from 3950 to 900 cm$^{-1}$, with spectral resolution of 4 cm$^{-1}$, transmission mode, and 256 and 64 background and sample co-added scans, respectively.

During the image acquisition, the mosaic region was positioned to cover each core, while also collecting the paraffin around the tissue. In addition, it was acquired a single mosaic using a clean slide and turning off the air purge of the microscope acrylic box to obtain spectra with water vapor (H2O) variation.

*2.2. Data preprocessing*

Images were preprocessed individually. Tissue, paraffin and possible pure slide regions were selected using a two K-means clustering in sequence. The first one clustered the raw spectra truncated at the Amide I and II region (1700 to 1500 cm$^{-1}$) into two clusters: tissue and paraffin + pure slide. Raw spectra were then truncated at the highest paraffin intensity band (1480 to 1450 cm$^{-1}$) and tissue previously clustered were set to zero for the second K-mean, grouping paraffin and zeroed tissue + pure slide. Spectra were preprocessed by the following steps:

Spectra were truncated in the biofingerprint region (1800 to 900 cm$^{-1}$), decreasing the size to 467 points. Outlier removal was performed by the Hotelling's $T^2$ vs Q residuals approach, with 10 Principal Components (PC) and removing spectra above the 95% confidence interval threshold. Spectra were smoothed adopting Savitzky-Golay method with window size of 11 and polynomial order of 2.

Extended Multiplicative Signal Correction (EMSC) [18] with digital de-waxing [19] and H2O removal was employed. PC quantity was selected until 99% of explained variance. Global mean spectrum, calculated from all samples, was used as reference. Baseline correction was accomplished by polynomial of order 4. EMSC model was solved by least squares estimation.

Corrected spectra were normalized by the min-max method, and a second outlier removal was applied. The 2D mosaics were reconstructed with preprocessed tissue spectra and zeroing paraffin and pure slide spectra. Final mosaics of 320x320x467 were divided into 6000 patches of 32x32x467. Patches with half or more of the pixel's quantity zeroed were excluded.

Patches were labeled by a binary encoding for the type and HER2 level (1+ and 2+ were not considered due to quantity limitations) classification; by a one-hot encoding for the subtype classification; and by an ordinal one-hot-like encoding [20] for the receptor levels. The percentage of Ki67 was min-max scaled to a 0 to 1 expression fraction for a regression.

*2.3. Deep learning*

A 2D convolutional neural network (CNN) called CaReNet-V2 was developed inspired on hyperspectral images classification [21–23], VGG [24], and generators of generative adversarial networks (GAN) [25,26]. **Fig. 1** presents the CaReNet-V2 architecture. The model has two channels path: one to target spectral feature extraction; and another for spatial extraction.

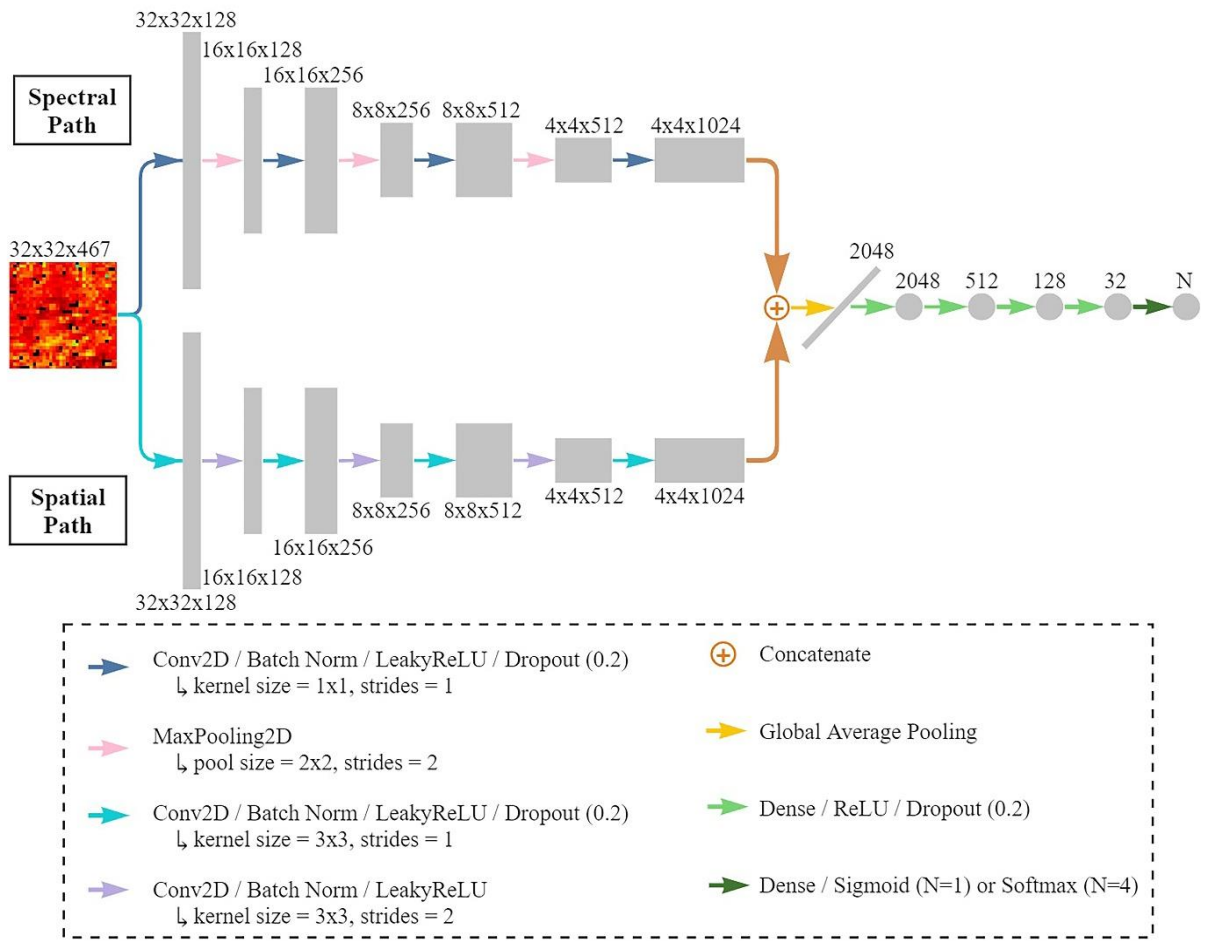

**Fig. 1.** CaReNet-V2 architecture.

Convolutional layers were created using HeNormal kernel initialization [27]. Zero padding was applied to both convolutional and pooling layers. A total of six models were created, one per label, where the final dense layer, activation and losses were dependent to the encoding: binary (type and HER2) – single neuron, sigmoid activation, and binary cross-entropy loss; one-hot encoded (subtype) – four neurons, softmax, and categorical cross-entropy; ordinal (ER and PR) – four neurons, sigmoid, and square error [20]; regression (Ki67) – one neuron, linear, and mean squared error.

Adam [28] was set as the optimization algorithm, with learning rate of 1e-3, beta 1 of 0.9 and beta 2 of 0.999. A cosine decay schedule with restarts [29] was applied with initial learning rate of 1e-3, first decay step with the length of the training set, epochs multiplier in the decay cycle (t_mul) of 1.5, initial learning rate multiplier (m_mul) of 1.0, and minimum learning rate (alpha) of 1e-5. Class weights were calculated and applied to the losses to correct the learning process with unbalanced classes.

Four patients were held-out for the test set, resulting in one patient of each class for multi-class and regression models, and two of each for binary models. The 26 remaining patients, or 21 remaining for Ki67 regression, were split in train and development (dev) sets by a stratified 4-fold cross-validation. Type, subtype and HER2 were split with unique patients for train and dev, while the others, due to quantity limitations, were split by patches, presenting a same patient in both train and dev sets.

Models were trained by a batch size of ten patches by 300 epochs. Training patches were randomly shuffled and augmented by each epoch using a data generator. Data augmentation involved different random rotation (90°, 180° and 270°) and flip (horizontally and vertically) transformations every epoch, without duplicating data or increasing the dataset size.

Performance was evaluated by each patch prediction and by the sample prediction with a voting system. Final classification predictions were standardized in relation to the encoding: a fixed threshold of 0.5 for binary approaches (type and HER2); the maximum argument for one-hot encoded (subtype); the maximum argument above 0.5 for ordinal encoding (ER and PR). Then, for the sample voting system, the most predicted class among all sample patches was chosen as the final classification, and Ki67 final regression was defined by the mean of all patches. Classification evaluation was accomplished by accuracy, specificity and sensitivity metrics, while the regression was assessed using the mean absolute error (MAE), mean squared error (MSE) and root-mean-square error (RMSE).

Gradient-weighted Class Activation Mapping (Grad-CAM) [30] was employed to analyze spatial activation. The best dev set model from each group was selected to calculated the Grad-CAM using the last convolutional layer of the spatial path channel. The channel importance, i.e., the contribution by wavenumber to the classification, was analyzed by summing the kernels values from the first convolutional layer of the spectral path. Spectral and spatial paths relative contributions were calculated by their respective sum of GAP feature values and first dense weights multiplication. All the study was performed by in house algorithms in Python, mainly Tensorflow and Keras libraries, and using a GeForce GTX 1080 GPU with 8 GB of memory.

## 3. Results and Discussion

**Fig. 2** depicts a representative image of the preprocessing process. The amide I peak demonstrates the impact of each process, as amide bands are indicators of biological tissue [10]. Paraffin blue border regions in Figure **Fig. 2 (a)** were clustered and zeroed by the K-means process in **Fig. 2 (b)**. Some residual tissue regions still appear after the clustering, evidenced by the blue chunks in the black zeroed paraffin area. This may be due to thin tissue residues, where the amide band intensities are present, but it is not thick enough to have a similar spectrum from the rest, thus being identified as outliers and eliminated after the remaining preprocessing steps in **Fig. 2 (c)**. Besides the clean black border, it is possible to visualize the more defined "holes" inside the core, where borderline outliers were excluded.

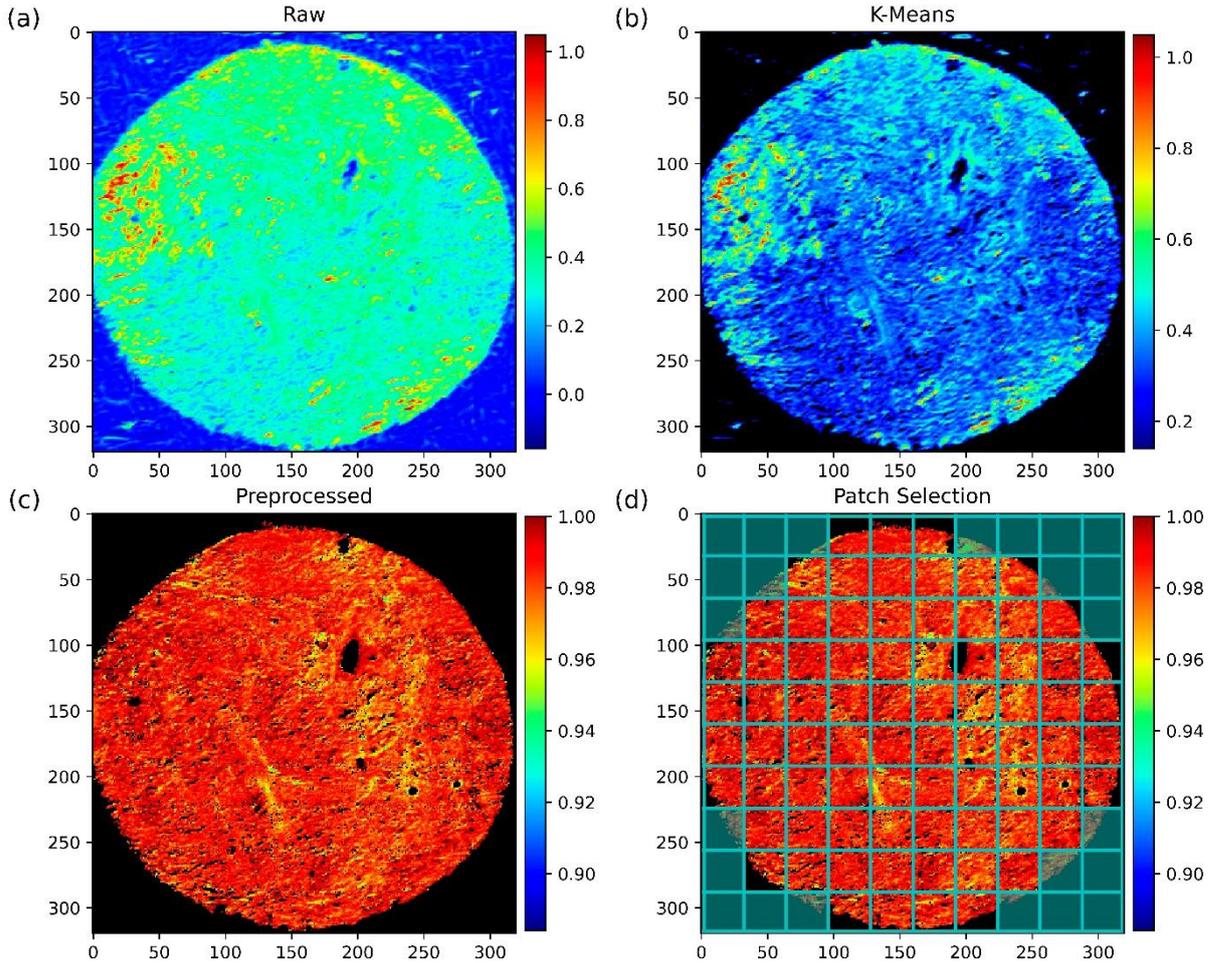

**Fig. 2.** Representative figure of the preprocessing process using the Amide I intensity peak image for better visualization. (a) Raw spectra; (b) Tissue raw spectra after K-means clustering. Paraffin as black (zeroed); (c) Preprocessed spectra; (d) Patches selection. Painted cyan squares are excluded patches. Spatial scale of images in pixels.

EMSC and normalization improved the scale presentation in **Fig. 2 (c)**, where values close to the maximum (1) are due to the fact that the amide I peak is usually the one with the highest intensity of the spectrum. The plot was normalized itself, hence improving the variation visualization. Patches with more than 50% of the pixels as zeroes were automatically removed, which were mainly the paraffin border patches, as shown in **Fig. 2 (d)**.

Test sets classifications performance are shown in **Table 2**. Type model presented the best metrics, as expected due to the disparity between malignant and benign tissues. Even though they were both above 0.9, the higher sensitivity than specificity, of 0.95 and 0.91, respectively, is preferred for cancer diagnosis, once the sensitivity is the probability of correct identifying a truly present cancer [31]. Therefore, it is better not to predict a false negative, while it is acceptable a certain level of false positives.

**Table 2**

CaReNet-V2 performance for test patches classification. Results grouped by classes of each model. Mean values ± standard deviation.

| Label | Class | Accuracy | Specificity | Sensitivity |
|---|---|---|---|---|
| Type | CA | 0.91 ± 0.02 | 0.91 ± 0.03 | 0.95 ± 0.02 |
| Subtype | LA | 0.85 ± 0.05 | 0.80 ± 0.06 | 0.89 ± 0.04 |
|  | LB | 0.84 ± 0.04 | 0.81 ± 0.05 | 0.77 ± 0.05 |
|  | HER2 | 0.90 ± 0.03 | 0.87 ± 0.04 | 0.91 ± 0.05 |
|  | TNBC | 0.92 ± 0.02 | 0.90 ± 0.03 | 0.89 ± 0.03 |
| ER | − | 0.65 ± 0.06 | 0.62 ± 0.07 | 0.69 ± 0.10 |
|  | + | 0.57 ± 0.12 | 0.62 ± 0.08 | 0.53 ± 0.09 |
|  | ++ | 0.59 ± 0.09 | 0.66 ± 0.10 | 0.53 ± 0.09 |
|  | +++ | 0.66 ± 0.07 | 0.61 ± 0.05 | 0.71 ± 0.06 |
| PR | − | 0.63 ± 0.05 | 0.58 ± 0.08 | 0.70 ± 0.08 |
|  | + | 0.54 ± 0.09 | 0.60 ± 0.08 | 0.51 ± 0.11 |
|  | ++ | 0.51 ± 0.07 | 0.56 ± 0.10 | 0.50 ± 0.12 |
|  | +++ | 0.62 ± 0.05 | 0.68 ± 0.10 | 0.68 ± 0.07 |
| HER2 | 3+ | 0.82 ± 0.04 | 0.84 ± 0.05 | 0.77 ± 0.06 |

Subtypes demonstrated metrics above 0.77, indicating a nice performance where the models learned how to extract features from the samples for this classification. HER2 and TNBC were better classified than LA and LB, with metrics around 0.9. This may be due the similarity between LA and LB samples, since their expression levels of receptors may be the same, demanding the Ki67 level to distinguish them [4].

ER and PR labels presented the lowest metrics, especially regarding borderline classes (+ and ++). The classification of the expression levels of each receptor is a finer prediction than the subtypes, since it is necessary to differentiate the same characteristics, at different levels, instead of a macro grouping of the characteristics as in the subtypes. Furthermore, the built dataset contains considerably fewer borderline classes samples (2 or 3) in comparison to others (10 or more samples), making generalization a difficult task, even with loss punishment by class weights.

HER2 models achieved metrics close to those of subtype, higher than ER and PR ones. However, a binary classification is easier than multi-class, especially in this case where the borderline classes (1+ and 2+) were not considered. Nevertheless, the models showed the ability of learning the HER2 evaluation.

Ki67 regression test performance is exhibited in **Table 3**. A mean difference between GT and predicted of 2.3% is a good overall prediction for the current dataset, 5 and 10% will remain as low level, and 20 and 30% as high level, since the cutoff is usually given by 15% [32]. MSE and RMSE measure the variance and standard deviation of the residuals, respectively, amplifying high errors more than lower ones. Therefore, the fact that they did not scale too far from the MAE indicates low impact from outliers.

**Table 3**

CaReNet-V2 performance for test patches Ki67 regression. MAE (Mean Absolute Error), MSE (Mean Squared Error) and RMSE (Root-Mean-Square Error) according to the predictions of the models on the min-max fraction scale (0-1 range) and rescaled to percentage (5-30% range). Mean values ± standard deviation.

| Scale | MAE | MSE | RMSE |
|---|---|---|---|
| Min-Max | 0.094 ± 0.015 | 0.021 ± 0.003 | 0.145 ± 0.009 |
| Rescaled (%) | 2.3 ± 0.4 | 13.1 ± 1.6 | 3.6 ± 0.2 |

Missing some patches predictions does not imply in a wrong final test patient prediction, as the voting system may overcome these mistakes. **Table 4** exhibits the final test patient classification results after the voting system.

**Table 4**

CaReNet-V2 classification performance for each test patient and each of the four models from the folds. Light blue corresponds to correct predictions in comparison to the GT (Ground Truth), whilst light red are wrong predictions.

| Label | GT Class | Predicted class – Model fold: 1 | 2 | 3 | 4 |
|---|---|---|---|---|---|
| Type | AT | AT | AT | AT | AT |
|  | AT | AT | AT | AT | AT |
|  | CA | CA | CA | CA | CA |
|  | CA | CA | CA | CA | CA |
| Subtype | LA | LA | LA | LA | LA |
|  | LB | LB | LA | LA | LB |
|  | HER2 | HER2 | HER2 | HER2 | HER2 |
|  | TNBC | TNBC | TNBC | TNBC | TNBC |
| ER | − | − | − | − | − |
|  | + | − | + | + | + |
|  | ++ | ++ | ++ | ++ | +++ |
|  | +++ | +++ | +++ | +++ | +++ |
| PR | − | − | − | − | − |
|  | + | + | + | − | + |
|  | ++ | − | ++ | ++ | +++ |
|  | +++ | +++ | +++ | +++ | +++ |
| HER2 | 0 | 0 | 0 | 0 | 0 |
|  | 0 | 0 | 0 | 0 | 0 |
|  | 3+ | 3+ | 0 | 3+ | 3+ |
|  | 3+ | 3+ | 3+ | 3+ | 0 |

The models from the four folds were able to correctly classify the Type (Cancer vs AT) for all test patients, withstanding the highest test patches metrics among all predictions. Type classification was expected to show the best performance, since it is a binary classification of the most different tissues: malignant and non-malignant. Still, a perfect cancer identification is an excellent characteristic for a screening tool.

Subtypes were classified correctly, except for two LB misclassified as LA. This also expresses the similarity between LA and LB, corroborating the findings in single patches metrics. It is important to discern luminal subtypes from HER2 and TNBC due to their higher incidence and better prognosis. LA represents 30 to 40% of breast cancers and LB 20 to 30%, while HER2 and TNBC ranges from 12 to 20% each [33,34].

Luminal subtypes demonstrates better treatment outcome and survival rate [35–37], whilst TNBC leads the worst, mainly because it develops resistance to its often treatment, chemotherapy, and has a high risk of evolving brain metastasis [35]. Luminal is also less recurrent, where LA evolves slowly within time and LB presents a peak incidence of recurrence in the first 5 years. On the other hand, HER2 and TNBC manifest a peak of recurrence in one or two year [33,38]. Hence, considering a screening technique, it is less critical to misclassify between LA and LA than a wrong HER2 or TNBC prediction.

ER and PR levels were satisfactory predicted, showing difficulties with borderline levels (+ and ++). This may be related to the lack of borderline samples, not providing enough examples for the models to learn their characteristics. Loss punishment by classes weights and the acquired knowledge from – and +++ samples may have assisted in the learning of these classes' prediction, however more samples are still necessary. If only – and +++ samples were considered, the models would be classifying mainly LA/LB vs HER2/TNBC, since there few examples of LA and LB with negative ER or PR. Performance could be improved if ER and PR expressions were grouped as negative (– and +) or positive (++ and +++) classification [39], once binary classifications are usually easier to be modeled. Despite this information being the most critical for the prognosis, these receptors play a substantial role in the assessment, and further details should be considered whenever possible.

Positive hormone receptor expression status is a favorable prognostic factor and a predictor of response to endocrine therapy [33]. Patients with both ER and PR positivity usually experience better outcomes than single positivity, especially single PR one, once believed a rare phenomenon, now reported as exhibiting a behavior as aggressive as HER2 and TNBC [40,41]. ER and PR positive and HER negative is the most prevalent with 60 to 70% of all breast cancers, where antiestrogen target therapy is associated with improvement in overall survival in both early and advanced phases, in addition to good responses to adjuvant chemotherapy [42]. Thereby, the proper analysis of these receptors can change the whole treatment strategy for a wide range of patients.

Instead of using four levels, receptors can be analyzed by the expression percentage [43,44]. It would be of great value to predict this percentage, using a regression process analogously to the Ki67 one, accomplished in this study. Nevertheless, this would require a whole new and larger dataset, with representative expressions from all the possible range assessed by gold standard techniques. This kind of samples should be evaluated in future studies.

HER2 levels were properly predicted, except for two 3+ classified as 0. Associated with the single patches' performance similar to the subtypes, which have four classes, this may indicate HER2 as a harder prediction, since it was a binary classification with larger samples number per class. Indeed, preliminary tests were performed using the 1+ and 2+ samples, in which the models were not able to learn this classification using only train/dev sets, as there were not enough samples for a test set, thus being omitted from this study. Even so, the ability to predict 0 or 3+ indicates that the model may learn the four levels classification if more samples examples is available.

Adequate HER2 assessment directly affects the prognosis. Tumors related to HER2-overexpression are regarded as aggressive neoplasms, associated with chemoresistance and poor survival rates. The most promising treatments are the use of tyrosine kinase inhibitors and immunotherapy with monoclonal antibodies [45]. These target therapies are usually employed in an adjuvant setting, and although had improved the prognosis, the high number of deaths from HER-positive breast cancers and researches for newer therapies persists [46,47].

**Table 5** displays Ki67 test patient predictions after the voting system. This is the only regression approach, hence not presenting the highlight for correct or not. Even though, the GT and predictions can be compared in terms of absolute error. Lower Ki67 levels presented better predictions than higher, with four-folds MAE of 1.5% for the lowest GT expression (5%), gradually increasing to 3.7% for the highest (30%).

**Table 5**

CaReNet-V2 Ki67 regression performance rescaled to percentage for each test patient and each of the four models from the folds. MAE (Mean Absolute Error) calculated from the four models' predictions with respect to the GT (Ground Truth).

| Label | GT % | Predicted % – Model fold: 1 | 2 | 3 | 4 | MAE % |
|---|---|---|---|---|---|---|
|  | 5 | 4.2 | 3.0 | 6.8 | 6.2 | 1.5 |
| Ki67 | 10 | 7.1 | 8.2 | 12.2 | 11.5 | 2.1 |
|  | 20 | 18.3 | 17.2 | 22.6 | 17.1 | 2.5 |
|  | 30 | 25.5 | 25.9 | 27.5 | 26.5 | 3.7 |

Using the MSE as the loss function helps dealing with outliers predictions. Decreasing outliers is important once the Ki67 is macro-divided in low or high expression if it is below or above 15%, respectively. Thus, near misses can still be in the same category, although outliers will probably lead to an incorrect one. Considering this cutoff point, all test patients were predicted within the correct low/high range, even though there was no sample with a borderline GT of 15% to better evaluate this occurrence. MSE usually does not deal well with imbalanced datasets [48], however the usage of class weights assisted to overcome this issue, as a large distribution difference was present on the dataset

Ki67 modeling involved four target values from two macro-levels, hence it could be more appropriated to deal with it as a classification approach. A binary low/high classification is useful, but it does not provide as much information as all the percentages, especially when dealing with borderline expressions. Even a categorical multi-class approach does not represent all possible real-life Ki67 levels. Therefore, the regression method was chosen to verify how the model would perform with an approach that could account all Ki67 expression levels, which can range from barely 0 to almost 100% [49]. Even so, it is required a much larger dataset with several samples in this range to properly evaluate this process.

Besides the usage to distinguish LA from HER2 negative LB, Ki67 expression is important to evaluate treatment responsiveness, endocrine or chemotherapy resistance, residual risk, and a dynamic biomarker during therapy [32]. High Ki67 is associated with poor survival, however the cutoff may vary between studies. It is reported variations on the cutoff of 10 to 20% [50,51]. Other assessments may also be indicated, such as relating a cutoff of 40% to a higher risk of recurrence and death for resected TNBC [52]. Therefore, a complete Ki67 regression is an advantageous analysis in comparison to binary or multi-class classifications.

**Fig. 3** depicts the Grad-CAM analysis. It is possible to visualize a well-distributed high intensity heatmap all over the tissue region, indicating the spatial contribution of a large area. Zeroed black pixels have a low classification contribution, as their weights are multiplied by zero. The spatial path of the model is responsible for spatial feature extraction by evaluating the 3x3 kernel, i.e., spatial evaluations of 9 spectra per step. Spectral path convolutions only assess individual pixels, where the downsampling is executed by pooling layers. Hence, the Grad-CAM of this path does not provide useful information, presenting meaningless heatmaps.

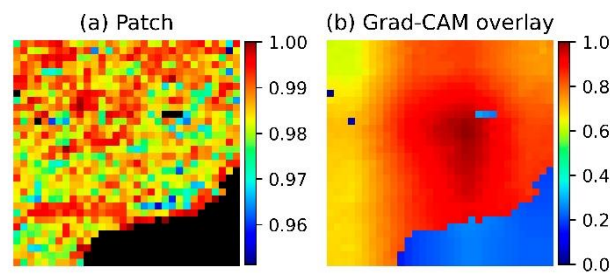

**Fig. 3**. Representative image of the Gradient-weighted Class Activation Mapping (Grad-CAM) for Type classification.

**Table 6** lists the most influential channels for each label prediction. These channels, or wavenumber bands, are assigned to biochemical information [53,54].

**Table 6**

Channel importance for each label. Top three wavenumber bands and their main assignments [53,54].

| Label | #1 Band (cm$^{-1}$) | Assignments | #2 Band (cm$^{-1}$) | Assignments | #3 Band (cm$^{-1}$) | Assignments |
|---|---|---|---|---|---|---|
| Type | 1658-1650 | Amide I | 1240-1236 | Phosphodiester $PO_2^-$ | 1553-1539 | Amide II |
| Subtype | 1597-1588 | Adenine Phenyl ring | 1753-1742 | Lipids Fatty Acids | 1073-1065 | Nucleic Acids Phosphate |
| ER | 1055-1050 | RNA DNA | 1612-1603 | Adenine | 1287-1281 | Collagen |
| PR | 1206-1192 | Collagen | 1047-1036 | Carbohydrates RNA | 1647-1641 | Amide I |
| HER2 | 972-968 | Nucleic acids | 1506-1493 | Amide II Phenyl rings | 1026-1017 | Glycogen |
| Ki67 | 1088-1085 | Phosphate $PO_2^-$ | 1242-1234 | Phosphodiester $PO_2^-$ | 1649-1639 | Amide I |

Amide I and II, listed in type, PR, HER2 and Ki67, have been used to differentiate cancerous and normal tissues [55], specially the 1655 cm$^{-1}$ in type importance range, which is related to α-helix amide I and is reported to have its intensity decreased for malignant tissues [53]. The 1240 cm$^{-1}$ in type and Ki67 is regarded to asymmetric non-hydrogen-bonded phosphate stretching modes from phosphodiester vibration, which suggest an increase in the nucleic acids in cancerous tissue.

Adenine bands, as in subtype and ER, are reported to be higher in patients with cancer. This is due to the higher accelerated metabolism of the cells, which entails in oscillatory deformations of the C–H peak of adenine [56]. Overproduction of fatty acids, listed in subtype importance, facilitates the tumor evolution and the survival of cancerous cells [56].

Differences in DNA and RNA vibration frequency, as in the band shown in #1 ER importance, is an important evaluation to discriminate between normal and cancer spectra [56]. PR importance presented the collagen (amide III) influence, where the 1204 $cm^{-1}$ is associated with higher intensities for breast carcinoma tissue [53]. Still in PR, it is noted an absence of carbohydrates peaks in breast cancer spectra, which may be related to the higher glucose metabolism in cancer cells [56].

Nucleic acids, as in HER2 #1 importance band, are augmented in cancer tissue due to their increase in the relative content [54]. The 1026 to 1017 $cm^{-1}$ HER2 #2 band, may be linked to the higher metabolism during the neoplastic process [57]. The 1086 $cm^{-1}$ listed in Ki67 band #1 due to symmetric stretching modes is reported to be increased in nucleic acids in malignant tissues [53].

Wavenumber shifts of the intensities may take place, where the band-region evaluation provides more reliable information. The channel importance analysis can help to understand the impact of the biochemical composition of the breast cancer to the predictions in a label-free approach. However, a fully 1D model may provide more information and facilitate interpretation, once the whole model extracts intensity features from single spectra, totally related to wavenumber variations and without any influence from neighboring spectra. In contrast, for 2D and 3D models it is necessary the evaluation of each channel individually, which becomes a difficult task since the models accounts the spatial relation of the spectra and their extracted features are mixed. Thus, only the first conv layer was considered for the current channel importance analysis, which is not directly related to the final model prediction, once the features pass through several other conv and dense layers.

The relative GAP importance analysis revealed that spectral path features accounted for 58 to 79% of the total contribution, while the rest was related to spatial features. This indicates the greater importance from extracting individual spectral features, with deeper assessment of the local biochemical information. Spatial path also calculates spectral features, although it takes more into account the spatial relationship of the spectra. Even so, it also plays an important role by analyzing heterogeneous samples such as breast cancer tissue.

Models created with a 2D approach demands more computing memory than 1D to be trained. CaReNet-V2 architecture with a 32x32x467 image patch as input and four neurons as output classification resulted in 15,914,660 parameters. On the other hand, predicting a patch implies 1024 spectra processed together, speeding up the overall prediction time for a large mosaic.

Other models reported on the literature were tested, such as the well-established VGG [24], and the state of the art models EfficientNet [58] and ConvNeXt [59]. Still, they were not able to learn how to extract features and properly classify the samples. Adding residuals convolutions increased the training time and did not improve the performance.

A 32x32x467 image can be compared to a deep hidden layer of a standard RGB (Red, Gree, Blue) classification model, where after several downsampling layers and progressive augmentation of the filters number to better extract the information, features maps images may approximate the size of the hyperspectral patches in this study [24,60]. In this way, for both spectral and spatial paths, the first convolutional layer with 128 filters assists to extract the most important information from the 467 wavenumber channels, decreasing the channels size at first, and then the next layers work on understanding these characteristics.

Patches approach was necessary to not overload the GPU memory, although it was also a benefit due to increasing the dataset size by 100x. Combined with the fact that hyperspectral imaging provides more information than RGB, it was possible to train the models with only 30 breast cancer patient samples. In addition, it enables the prediction of any mosaic size as long as it is possible to build 32x32 patches. Voting system assists for a better prediction, since it is not dependent on a single patch evaluation, thus final prediction may overcome possible mistakes.

All spectral path layers containing 1x1 kernels causes each spectrum features to be extracted individually, with no influences of neighbors' spectra. The max pooling layers select the most discriminating features, without reprocessing them with convolution kernels. It would be possible to keep extracting features with 1x1 kernels and apply a single Global Max Pooling (GMP) at the end of the spectral path, instead of the GAP, but that would make the model more complex and remove the most discriminating feature emphasizing. GAP before dense layers is preferred to map the complete extent of extracted features instead of selecting the most discriminative ones as in GMP, besides acting as a structural regularizer and helping to prevent overfitting [61].

Loss punishment based on class weights supports the development of imbalanced datasets without using oversampling or downsampling approaches. Nevertheless, the training process with balanced classes, avoiding class weights or other balancing methods, usually present better results [62]. Thereupon, a larger and balanced dataset may considerably improve CaReNet-V2 performance. Increasing dataset size would also provide more test samples, aiding to achieve a better real world performance evaluation.

Other tested approaches did not exhibit satisfactory results and were omitted from this study for simplicity. Even so, it is worth describing them to guide future research: second derivative led to dummy classifiers; PCA instead of patches, downsampling the input to 320x320x10, did not supply enough information for the feature extraction; reduce learning rate on plateau callback [63] was not able to leave local minima; one-hot rather than ordinal encoding for ER and PR levels displayed lower metrics, possibly due to the models not learning the ranking relationship between the levels.

A microarray sample with representative subtypes distribution was chosen for this study as subtypes stratifies patients for treatment, guiding the systemic therapy in preoperative, postoperative or both scenarios [34,64]. Yet, a complete pathology report screening tool should address other evaluation methods, such as breast cancer histology (50 to 75% of patients present invasive ductal carcinoma while 5 to 15% shows invasive lobular carcinoma) [34], TNM (Tumor, Node, Metastasis) staging [65], and grade system [66], once these methods may also assist to a better understanding of the breast cancer and should be considered in future studies with larger samples number.

## 4. Conclusion

The fully automated preprocessing approach allows for user-independent data preparation. Non-biological tissue regions were excluded from the images, and patches selection provided an increase in the dataset size, where one mosaic may result in 100 patches. This enabled to develop and train the models with only 30 breast cancer patients.

CaReNet-V2 showed a perfect adjacent tissue (AT) vs breast cancer tissue (CA) classification and only two Luminal A/Luminal B mistaken for test patients' classification. Furthermore, it provided the biomarkers (estrogen receptor – ER, progesterone receptor – PR, human epidermal growth factor receptor 2 – HER2, and Ki67) levels prediction with good overall metrics, but demonstrated lower performance for borderline classes.

Future studies should consider more samples to achieve other evaluations of the pathology report and possibly improve metrics and generalization of the model. The novel 2D deep learning and hyperspectral imaging approach in this study is a potential technique for breast cancer biopsies evaluation, standing out as a screening analysis technique and helping to prioritize patients.

**Acknowledgements**

The authors would like to thank the Health Innovation Techcenter of Hospital Israelita Albert Einstein for sharing their GPU workstation.

**Conflict of interest**

The authors do not have any conflicts of interest to disclose.

**Funding**

This work was supported by FAPESP (CEPID 05/51689-2, 17/50332-0), CAPES (Finance Code 001, PROCAD 88881.068505/2014-01) and CNPq (INCT-465763/2014-6, PQ-309902/2017-7, 142229/2019-9).

**References**

[1] Sung H, Ferlay J, Siegel RL, Laversanne M, Soerjomataram I, Jemal A, et al. Global Cancer Statistics 2020: GLOBOCAN Estimates of Incidence and Mortality Worldwide for 36 Cancers in 185 Countries. CA Cancer J Clin 2021;71:209–49. https://doi.org/10.3322/caac.21660.

[2] Hennigs A, Riedel F, Gondos A, Sinn P, Schirmacher P, Marmé F, et al. Prognosis of breast cancer molecular subtypes in routine clinical care: A large prospective cohort study. BMC Cancer 2016;16:734. https://doi.org/10.1186/s12885-016-2766-3.

[3] Goldhirsch A, Wood WC, Coates AS, Gelber RD, Thürlimann B, Senn HJ. Strategies for subtypes-dealing with the diversity of breast cancer: Highlights of the St Gallen international expert consensus on the primary therapy of early breast cancer 2011. Ann Oncol 2011;22:1736–47. https://doi.org/10.1093/annonc/mdr304.

[4] Kos Z, Dabbs DJ. Biomarker assessment and molecular testing for prognostication in breast cancer. Histopathology 2016;68:70–85. https://doi.org/10.1111/his.12795.

[5] Russnes HG, Lingjærde OC, Børresen-Dale AL, Caldas C. Breast Cancer Molecular Stratification: From Intrinsic Subtypes to Integrative Clusters. Am J Pathol

2017;187:2152–62. https://doi.org/10.1016/j.ajpath.2017.04.022.

[6] Zhu J, Liu M, Li X. Progress on deep learning in digital pathology of breast cancer: a narrative review. Gland Surg 2022;11:751–66. https://doi.org/10.21037/gs-22-11.

[7] Kalmodia S, Parameswaran S, Yang W, Barrow CJ, Krishnakumar S. Attenuated total reflectance Fourier Transform Infrared spectroscopy: an analytical technique to understand therapeutic responses at the molecular level. Sci Rep 2015;5:16649. https://doi.org/10.1038/srep16649.

[8] Kumar S, Srinivasan A, Nikolajeff F. Role of Infrared Spectroscopy and Imaging in Cancer Diagnosis. Curr Med Chem 2017;25:1055–72. https://doi.org/10.2174/0929867324666170523121314.

[9] Su KY, Lee WL. Fourier transform infrared spectroscopy as a cancer screening and diagnostic tool: A review and prospects. Cancers (Basel) 2020;12:115. https://doi.org/10.3390/cancers12010115.

[10] Gautam R, Vanga S, Ariese F, Umapathy S. Review of multidimensional data processing approaches for Raman and infrared spectroscopy. EPJ Tech Instrum 2015;2:8. https://doi.org/10.1140/epjti/s40485-015-0018-6.

[11] Lecun Y, Bengio Y, Hinton G. Deep learning. Nature 2015;521:436–44. https://doi.org/10.1038/nature14539.

[12] Giambagli L, Buffoni L, Carletti T, Nocentini W, Fanelli D. Machine learning in spectral domain. Nat Commun 2021;12:1330. https://doi.org/10.1038/s41467-021-21481-0.

[13] Wang L, Ren B, He H, Yan S, Lyu D, Xu M, et al. Deep learning for biospectroscopy and biospectral imaging: State-of-the-art and perspectives. Anal Chem 2021;93:3653–65. https://doi.org/10.1021/acs.analchem.0c04671.

[14] Yang J, Xu J, Zhang X, Wu C, Lin T, Ying Y. Deep learning for vibrational spectral analysis: Recent progress and a practical guide. Anal Chim Acta 2019;1081:6–17. https://doi.org/10.1016/j.aca.2019.06.012.

[15] Tomas RC, Sayat AJ, Atienza AN, Danganan JL, Ramos MR, Fellizar A, et al. Detection

of breast cancer by ATR-FTIR spectroscopy using artificial neural networks. PLoS One 2022;17:e0262489. https://doi.org/10.1371/journal.pone.0262489.

[16] Du Y, Xie F, Yin L, Yang Y, Yang H, Wu G, et al. Breast cancer early detection by using Fourier-transform infrared spectroscopy combined with different classification algorithms. Spectrochim Acta Part A Mol Biomol Spectrosc 2022;283:121715. https://doi.org/10.1016/j.saa.2022.121715.

[17] Berisha S, Lotfollahi M, Jahanipour J, Gurcan I, Walsh M, Bhargava R, et al. Deep learning for FTIR histology: leveraging spatial and spectral features with convolutional neural networks. Analyst 2019;144:1642–53. https://doi.org/10.1039/c8an01495g.

[18] Afseth NK, Kohler A. Extended multiplicative signal correction in vibrational spectroscopy, a tutorial. Chemom Intell Lab Syst 2012;117:92–9. https://doi.org/10.1016/j.chemolab.2012.03.004.

[19] De Lima FA, Gobinet C, Sockalingum G, Garcia SB, Manfait M, Untereiner V, et al. Digital de-waxing on FTIR images. Analyst 2017;142:1358–70. https://doi.org/10.1039/c6an01975g.

[20] Jianlin Cheng, Zheng Wang, Pollastri G. A neural network approach to ordinal regression. 2008 IEEE Int. Jt. Conf. Neural Networks (IEEE World Congr. Comput. Intell., IEEE; 2008, p. 1279–84. https://doi.org/10.1109/IJCNN.2008.4633963.

[21] Xu Y, Li Z, Li W, Du Q, Liu C, Fang Z, et al. Dual-Channel Residual Network for Hyperspectral Image Classification With Noisy Labels. IEEE Trans Geosci Remote Sens 2022;60:1–11. https://doi.org/10.1109/TGRS.2021.3057689.

[22] Yu S, Jia S, Xu C. Convolutional neural networks for hyperspectral image classification. Neurocomputing 2017;219:88–98. https://doi.org/10.1016/j.neucom.2016.09.010.

[23] Song H, Yang W, Dai S, Du L, Sun Y. Using dual-channel CNN to classify hyperspectral image based on spatial-spectral information. Math Biosci Eng 2020;17:3450–77. https://doi.org/10.3934/mbe.2020195.

[24] Simonyan K, Zisserman A. Very deep convolutional networks for large-scale image

recognition. 3rd Int. Conf. Learn. Represent. ICLR 2015 - Conf. Track Proc., 2015.

[25] Creswell A, White T, Dumoulin V, Arulkumaran K, Sengupta B, Bharath AA. Generative Adversarial Networks: An Overview. IEEE Signal Process Mag 2018;35:53–65. https://doi.org/10.1109/MSP.2017.2765202.

[26] Iqbal T, Ali H. Generative Adversarial Network for Medical Images (MI-GAN). J Med Syst 2018;42:231. https://doi.org/10.1007/s10916-018-1072-9.

[27] He K, Zhang X, Ren S, Sun J. Delving Deep into Rectifiers: Surpassing Human-Level Performance on ImageNet Classification 2015.

[28] Kingma DP, Ba J. Adam: A Method for Stochastic Optimization 2014.

[29] Loshchilov I, Hutter F. SGDR: Stochastic Gradient Descent with Warm Restarts 2016.

[30] Selvaraju RR, Cogswell M, Das A, Vedantam R, Parikh D, Batra D. Grad-CAM: Visual Explanations from Deep Networks via Gradient-Based Localization. Int J Comput Vis 2020;128:336–59. https://doi.org/10.1007/s11263-019-01228-7.

[31] Herman CR, Gill HK, Eng J, Fajardo LL. Screening for Preclinical Disease: Test and Disease Characteristics. Am J Roentgenol 2002;179:825–31. https://doi.org/10.2214/ajr.179.4.1790825.

[32] Nahed AS, Shaimaa MY. Ki-67 as a prognostic marker according to breast cancer molecular subtype. Cancer Biol Med 2016;13:496. https://doi.org/10.20892/j.issn.2095-3941.2016.0066.

[33] Fragomeni SM, Sciallis A, Jeruss JS. Molecular Subtypes and Local-Regional Control of Breast Cancer. Surg Oncol Clin N Am 2018;27:95–120. https://doi.org/10.1016/j.soc.2017.08.005.

[34] Waks AG, Winer EP. Breast Cancer Treatment. JAMA 2019;321:288. https://doi.org/10.1001/jama.2018.19323.

[35] Lv Y, Ma X, Du Y, Feng J. Understanding Patterns of Brain Metastasis in Triple-Negative Breast Cancer and Exploring Potential Therapeutic Targets. Onco Targets Ther 2021;Volume 14:589–607. https://doi.org/10.2147/OTT.S293685.

[36] Hwang K-T, Kim J, Jung J, Chang JH, Chai YJ, Oh SW, et al. Impact of Breast Cancer Subtypes on Prognosis of Women with Operable Invasive Breast Cancer: A Population-based Study Using SEER Database. Clin Cancer Res 2019;25:1970–9. https://doi.org/10.1158/1078-0432.CCR-18-2782.

[37] Howlader N, Cronin KA, Kurian AW, Andridge R. Differences in Breast Cancer Survival by Molecular Subtypes in the United States. Cancer Epidemiol Biomarkers Prev 2018;27:619–26. https://doi.org/10.1158/1055-9965.EPI-17-0627.

[38] Tsoutsou PG, Vozenin M-C, Durham A-D, Bourhis J. How could breast cancer molecular features contribute to locoregional treatment decision making? Crit Rev Oncol Hematol 2017;110:43–8. https://doi.org/10.1016/j.critrevonc.2016.12.006.

[39] Yang F, Li J, Zhang H, Zhang S, Ye J, Cheng Y, et al. Correlation between Androgen Receptor Expression in Luminal B (HER–2 Negative) Breast Cancer and Disease Outcomes. J Pers Med 2022;12:1988. https://doi.org/10.3390/jpm12121988.

[40] Zhao H, Gong Y. The Prognosis of Single Hormone Receptor-Positive Breast Cancer Stratified by HER2 Status. Front Oncol 2021;11. https://doi.org/10.3389/fonc.2021.643956.

[41] Fan Y, Ding X, Xu B, Ma F, Yuan P, Wang J, et al. Prognostic Significance of Single Progesterone Receptor Positivity. Medicine (Baltimore) 2015;94:e2066. https://doi.org/10.1097/MD.0000000000002066.

[42] Andrahennadi S, Sami A, Manna M, Pauls M, Ahmed S. Current Landscape of Targeted Therapy in Hormone Receptor-Positive and HER2-Negative Breast Cancer. Curr Oncol 2021;28:1803–22. https://doi.org/10.3390/curroncol28030168.

[43] Allison KH, Hammond MEH, Dowsett M, McKernin SE, Carey LA, Fitzgibbons PL, et al. Estrogen and Progesterone Receptor Testing in Breast Cancer: ASCO/CAP Guideline Update. J Clin Oncol 2020;38:1346–66. https://doi.org/10.1200/JCO.19.02309.

[44] Yi M, Huo L, Koenig KB, Mittendorf EA, Meric-Bernstam F, Kuerer HM, et al. Which threshold for ER positivity? a retrospective study based on 9639 patients. Ann Oncol

2014;25:1004–11. https://doi.org/10.1093/annonc/mdu053.

[45] English DP, Roque DM, Santin AD. HER2 Expression Beyond Breast Cancer: Therapeutic Implications for Gynecologic Malignancies. Mol Diagn Ther 2013;17:85–99. https://doi.org/10.1007/s40291-013-0024-9.

[46] Figueroa-Magalhães MC, Jelovac D, Connolly RM, Wolff AC. Treatment of HER2-positive breast cancer. The Breast 2014;23:128–36. https://doi.org/10.1016/j.breast.2013.11.011.

[47] Loibl S, Gianni L. HER2-positive breast cancer. Lancet 2017;389:2415–29. https://doi.org/10.1016/S0140-6736(16)32417-5.

[48] Wang S, Liu W, Wu J, Cao L, Meng Q, Kennedy PJ. Training deep neural networks on imbalanced data sets. 2016 Int. Jt. Conf. Neural Networks, IEEE; 2016, p. 4368–74. https://doi.org/10.1109/IJCNN.2016.7727770.

[49] Liang Q, Ma D, Gao R-F, Yu K-D. Effect of Ki-67 Expression Levels and Histological Grade on Breast Cancer Early Relapse in Patients with Different Immunohistochemical-based Subtypes. Sci Rep 2020;10:7648. https://doi.org/10.1038/s41598-020-64523-1.

[50] Choi SB, Park JM, Ahn JH, Go J, Kim J, Park HS, et al. Ki-67 and breast cancer prognosis: does it matter if Ki-67 level is examined using preoperative biopsy or postoperative specimen? Breast Cancer Res Treat 2022;192:343–52. https://doi.org/10.1007/s10549-022-06519-1.

[51] Skjervold AH, Pettersen HS, Valla M, Opdahl S, Bofin AM. Visual and digital assessment of Ki-67 in breast cancer tissue - a comparison of methods. Diagn Pathol 2022;17:45. https://doi.org/10.1186/s13000-022-01225-4.

[52] Wu Q, Ma G, Deng Y, Luo W, Zhao Y, Li W, et al. Prognostic Value of Ki-67 in Patients With Resected Triple-Negative Breast Cancer: A Meta-Analysis. Front Oncol 2019;9. https://doi.org/10.3389/fonc.2019.01068.

[53] Movasaghi Z, Rehman S, ur Rehman DI. Fourier Transform Infrared (FTIR) Spectroscopy of Biological Tissues. Appl Spectrosc Rev 2008;43:134–79.

https://doi.org/10.1080/05704920701829043.

[54] Malek K, Wood BR, Bambery KR. FTIR Imaging of Tissues: Techniques and Methods of Analysis, 2014, p. 419–73. https://doi.org/10.1007/978-94-007-7832-0_15.

[55] Kar S, Katti DR, Katti KS. Fourier transform infrared spectroscopy based spectral biomarkers of metastasized breast cancer progression. Spectrochim Acta Part A Mol Biomol Spectrosc 2019;208:85–96. https://doi.org/10.1016/j.saa.2018.09.052.

[56] Kołodziej M, Kaznowska E, Paszek S, Cebulski J, Barnaś E, Cholewa M, et al. Characterisation of breast cancer molecular signature and treatment assessment with vibrational spectroscopy and chemometric approach. PLoS One 2022;17:e0264347. https://doi.org/10.1371/journal.pone.0264347.

[57] Cappelletti V, Iorio E, Miodini P, Silvestri M, Dugo M, Daidone MG. Metabolic Footprints and Molecular Subtypes in Breast Cancer. Dis Markers 2017;2017:1–19. https://doi.org/10.1155/2017/7687851.

[58] Tan M, Le Q V. EfficientNet: Rethinking Model Scaling for Convolutional Neural Networks 2019.

[59] Liu Z, Mao H, Wu C-Y, Feichtenhofer C, Darrell T, Xie S. A ConvNet for the 2020s 2022.

[60] Ahmed WS, Karim A amir A. The Impact of Filter Size and Number of Filters on Classification Accuracy in CNN. 2020 Int. Conf. Comput. Sci. Softw. Eng., IEEE; 2020, p. 88–93. https://doi.org/10.1109/CSASE48920.2020.9142089.

[61] Lin M, Chen Q, Yan S. Network In Network 2013.

[62] Dablain D, Jacobson KN, Bellinger C, Roberts M, Chawla N. Understanding CNN Fragility When Learning With Imbalanced Data 2022.

[63] Smith LN, Topin N. Super-Convergence: Very Fast Training of Neural Networks Using Large Learning Rates 2017.

[64] Wolf DM, Yau C, Wulfkuhle J, Brown-Swigart L, Gallagher RI, Lee PRE, et al. Redefining breast cancer subtypes to guide treatment prioritization and maximize

response: Predictive biomarkers across 10 cancer therapies. Cancer Cell 2022;40:609-623.e6. https://doi.org/10.1016/j.ccell.2022.05.005.

[65] Hortobagyi GN, Edge SB, Giuliano A. New and Important Changes in the TNM Staging System for Breast Cancer. Am Soc Clin Oncol Educ B 2018:457–67. https://doi.org/10.1200/EDBK_201313.

[66] van Dooijeweert C, Baas IO, Deckers IAG, Siesling S, van Diest PJ, van der Wall E. The increasing importance of histologic grading in tailoring adjuvant systemic therapy in 30,843 breast cancer patients. Breast Cancer Res Treat 2021;187:577–86. https://doi.org/10.1007/s10549-021-06098-7.